\documentclass[10pt,twocolumn,letterpaper]{article}

\usepackage[pagenumbers]{wacv} 
\usepackage{algorithm}
\usepackage{algpseudocode}
\usepackage{multirow}
\usepackage{multicol}
\usepackage{listings}
\usepackage{makecell}
\usepackage{tabularx}
\usepackage[hyphens]{url}

\definecolor{wacvblue}{rgb}{0.21,0.49,0.74}
\usepackage[pagebackref,breaklinks,colorlinks,allcolors=wacvblue]{hyperref}
\def\wacvPaperID{2769} 
\def\confName{WACV}
\def\confYear{2026}

\title{Evaluating the Capability of Video Question Generation for Expert Knowledge Elicitation}

\author{
Huaying Zhang$^{1,2}$,
Atsushi Hashimoto$^{1}$,
Tosho Hirasawa$^{1}$
\\
$^{1}$OMRON SINIC X Corp., Japan,
$^{2}$Hokkaido University, Japan\\
{\tt\small huaying@lmd.ist.hokudai.ac.jp, \{atsushi.hashimoto,tosho.hirasawa\}@sinicx.com}
}

\begin{document}
\maketitle
\begin{abstract}
Skilled human interviewers can extract valuable information from experts. This raises a fundamental question: what makes some questions more effective than others? To address this, a quantitative evaluation of question‑generation models is essential.
Video question generation (VQG) is a topic for video question answering (VideoQA), where questions are generated for given answers.
Their evaluation typically focuses on the ability to answer questions, rather than the quality of generated questions.
In contrast, we focus on the question quality in eliciting unseen knowledge from human experts.
For a continuous improvement of VQG models, we propose a protocol that evaluates the ability by simulating question-answering communication with experts using a question-to-answer retrieval.
We obtain the retriever by constructing a novel dataset, EgoExoAsk, which comprises 27,666 QA pairs generated from Ego-Exo4D's expert commentary annotation.
The EgoExoAsk training set is used to obtain the retriever, and the benchmark is constructed on the validation set with Ego-Exo4D video segments.
Experimental results demonstrate our metric reasonably aligns with question generation settings: models accessing richer context are evaluated better, supporting that our protocol works as intended. The EgoExoAsk dataset is available in \href{https://github.com/omron-sinicx/VQG4ExpertKnowledge}{our project page}.

\end{abstract}

\section{Introduction}
Knowledge sharing is vital for societal development, as demonstrated by advances in the internet and large language models (LLMs). This progress is now extending into new fields, such as vision-language-action models \cite{rtx2023,shirai2024vilain,vla-survey2025}, opening possibilities to address various societal challenges. However, with internet resources nearing their limits~\cite{runout_2024}, the need to articulate and store knowledge efficiently becomes increasingly critical~\cite{li2025survey,video_label_alignment}. 

Consulting human experts offers a potential way to support knowledge elicitation, as recent multimodal LLMs (MLLMs) can lower questioning costs.
For human experts, effective questions are crucial to reduce response costs.
However, there is no established metric to assess question quality.
Quantitative evaluation is essential for analyzing and continuously improving the skills behind high-quality question generation.
This study aims to develop a reliable evaluation framework for the video question generation (VQG) task based on expert commentaries from the Ego-Exo4D dataset~\cite{egoexo4d}.

\begin{figure}[t]
  \centering
  \includegraphics[width=1.0\linewidth]{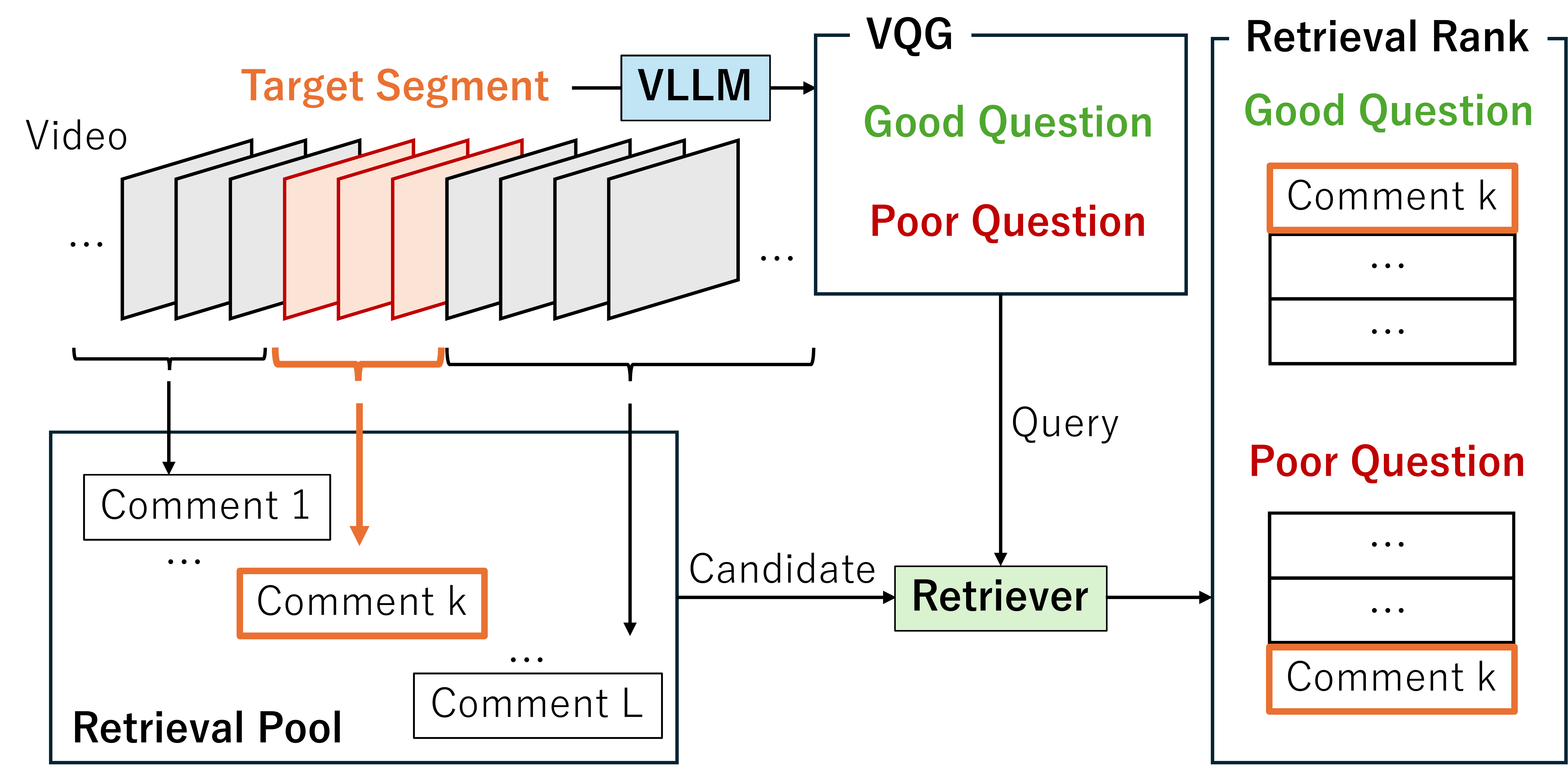}
   \caption{Our evaluation framework. A good question leads to segment-specific commentary even without knowing the answer. The QA retriever is trained to emulate this type of communication. Higher retrieval scores for generated questions indicate more effective question generation.}
   \label{fig:eyecatch}
\end{figure}

What makes a video question better than others? From the perspective of expert-knowledge elicitation, it should elicit expert commentary specific to the target video segment.
Based on this principle, we developed our retrieval-based evaluation protocol, illustrated in Fig. \ref{fig:eyecatch}.
This protocol uses a question-answer (QA) retriever, which identifies the best matching answer to the given question. This approximates the process by which an expert recalls an answer to a question. 
This metric judges a video question generator as superior when its questions retrieve expert commentary answers via the QA retriever better than other VQG models.

The protocol requires the core module, QA retriever. We obtain this by converting Ego-Exo4D's expert commentary dataset. The dataset provides one or multiple comments from experts for each event in a video. Since their format is commentary (e.g., {\it ``This climber is showing... an understanding of what that means by brushing the hold...''}\footnote{Actual sample from Ego-Exo4D's expert commentary dataset.}), we asked an LLM to convert them into QA form (e.g., Q: {\it ``Why might brushing the climbing holds before grasping them be an intentional choice during a climb?''} and A: {\it ``...this action indicates an awareness of improving grip quality...''}\footnote{Converted example in EgoExoAsk}). We call this the EgoExoAsk dataset. Based on the EgoExoAsk dataset (excluding the validation set), we trained the QA retriever.

In the experiment, we prepared multiple settings of VQG, which can access different information resources for question generation. Our metric reasonably evaluates and sorts their performance based on resource richness.
The contribution of this paper is threefold:
\begin{enumerate}
    \item We propose a metric that evaluates the performance of video question generators for the first time.
    \item We propose a novel benchmark protocol for the video question generation task, which incorporates the EgoExoAsk dataset.
    \item We demonstrate that the proposed metric reasonably evaluates a series of video question generating settings with multiple multi-modal LLMs.
\end{enumerate}

\section{Related Work}
\subsection{Question Generation in Visual QA Datasets}
In VideoQA and ImageQA, question generation often relies on structured or rule-based methods to evaluate visual-language models' answering capabilities. Traditional approaches~\cite{zhu2016visual7w,johnson2017clevr,lei2018tvqa}, use predefined templates to form questions by mapping visual content into structured formats. These methods, while systematic, lack adaptability and often fail to capture the complexity of real-world scenarios, limiting the depth of inquiry into implicit content and expert-level understanding. A recently proposed dataset, HD-EPIC~\cite{perrett2025hd}, provides more fine-grained questions. However, this dataset is restricted to the cooking video domain, and huge amounts of careful annotations are needed to create these questions. To mitigate the dependency on templates, researchers seek to generate QA datasets~\cite{jang2017tgif,xu2017video} automatically, but struggle with the quality. Recently, thanks to the explosive growth of LLMs, the quality of generated questions has been impressively improved. For example, EgoSchema~\cite{mangalam2023egoschema} and MM-Ego~\cite{ye2025mmego} both proposed a VideoQA dataset by using LLMs to generate questions from dense captions. Unlike these methods focused on the visible knowledge, the question generation task we discuss in this paper aims to extract the invisible knowledge behind the observed scene.

\subsection{Generated Question Assessment}
When evaluating the quality of generated questions, human assessment is commonly used~\cite{lei2018tvqa, Uehara_2023_WACV}. However, the assessment mainly targets dataset quality, not the method used to obtain the questions. In addition, human assessment requires significant costs and thus is not suitable as a general evaluation metric for VQG models. To reduce effort and cost, LLM-as-a-judge~\cite{gu2024survey, wang2024self} has been widely researched for evaluation across domains. On the other hand, the hallucination remains a concern for reliable evaluation~\cite{huang2025survey}. Although a study in the education field attempted to use LLMs for question evaluation, the results were not optimal~\cite{moore2023assessing}. Therefore, in this paper, we do not directly use the LLMs to score question quality. Some researches assess the question quality by training a video-language model on the generated QA pairs and measuring gains on comprehensive video-language tasks~\cite{yang2021just,wang2024vigc}, which is not the focus of this paper. \cite{xie2025explicitly} studies the method of generating questions at different levels of difficulty, but its focus is on constructing the VQG models rather than evaluation.

\section{Preliminaries}
\textbf{Traditional VQG.}
Given a video segment $s$, we assume that it contains knowledge that is easy to observe for non-experts (noted as $\mathcal{S}$) and invisible expert knowledge (noted as $\mathcal{S}^\prime$). Traditional VQG focuses on generating the question $q$ whose answer $a$ is obvious in the segment $s$.
\begin{equation}
    q=\mathcal{M}(a,s),a\in\mathcal{S},
\end{equation}
where $\mathcal{M}$ is the question generation model.

\textbf{VQG for Expert Knowledge Elicitation.}
Unlike traditional VQG, our task requires generating questions $q$ from $s$ without $a$, as $a$ is the expert knowledge inaccessible to $\mathcal{M}$. As a typical scenario, we focus on videos of people conducting specific tasks (i.e., playing sports, conducting experiments, etc.). Each segment $s$ of those videos reflects a component operation of the task. This problem is formulated as:
\begin{equation}
    q=\mathcal{M}(s), \label{eq:our_task}
\end{equation}
where the answer to $q$ should be $a\in \mathcal{S}^\prime$.


\textbf{Retrieval-based Question Evaluation.}
The main challenge to evaluate $\mathcal{M}$ in Eq. \eqref{eq:our_task} lies in the absence of ground truth $q^*$.
Hence, we propose a protocol that evaluates $q$ indirectly through its ability to elicit expert knowledge in $\mathcal{S}^\prime$. Specifically, we approximate the ``elicitation'' to ``retrieval'', i.e., we define the question as ``good'' if it can retrieve the positive comments specific to the video segment. For each segment, we construct a retrieval pool $\mathcal{R}$ consisting of these positive comments and negative comments of a similar situation (i.e., comments from a different segment of the same video) to ensure the difficulty of evaluation.

\textbf{Retriever.}
To conduct the retrieval, we use an embedding model $\mathcal{E}(\cdot)$ to calculate the similarity between the question and comments. To realize optimal retrieval, the embedding model, used as a retriever, should be more sensitive to the expert-knowledge domain than general models. That is to say, for an ideal question $q$, a positive comment $c^P\in\mathcal{R}$ related to the video segment $s$, and a negative comment $c^N\in\mathcal{R}$ not related to $s$, the retriever is expected to perform as follows:
\begin{equation}
    sim(\mathcal{E}(q), \mathcal{E}(c^P)) > sim(\mathcal{E}(q), \mathcal{E}(c^N)),
\end{equation}
where $sim(\cdot)$ represents cosine similarity.


\section{Retriever for Expert Knowledge Elicitation Evaluation}
\label{sec4}
In this section, we will introduce the method of obtaining the retriever $\mathcal{E}$ for evaluating the VQG for expert knowledge elicitation. We describe the process of the construction of our EgoExoAsk dataset in Sec.~\ref{subsec:egoexoask} and the retriever training using this dataset in Sec.~\ref{subsec:retriever}.

\begin{figure*}[t]
  \centering
  \includegraphics[width=\linewidth]{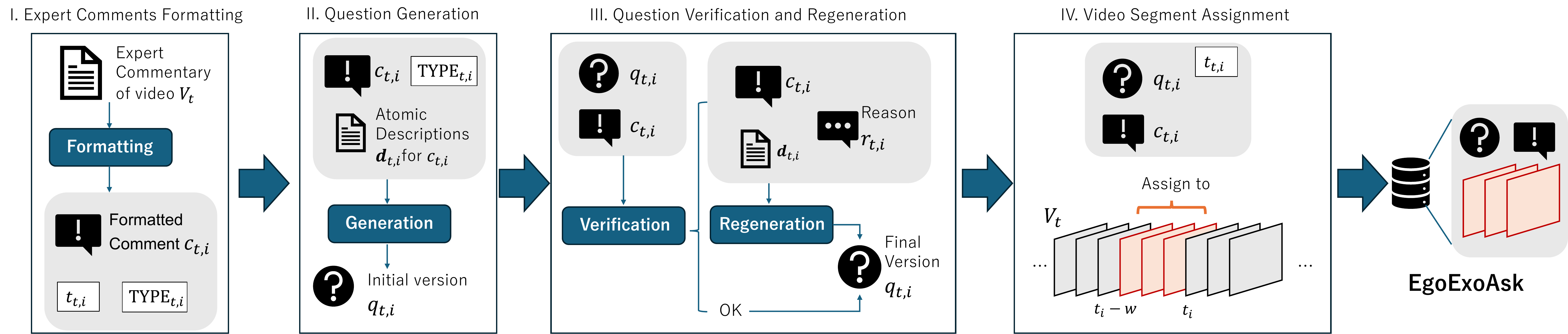}
   \caption{The pipeline for constructing our EgoExoAsk dataset. The components in blue are LLMs.}
   \label{fig:egoexoask}
\end{figure*}

\subsection{The EgoExoAsk dataset}
\label{subsec:egoexoask}
We use Ego-Exo4D dataset~\cite{egoexo4d} to develop the benchmark for the expert knowledge elicitation task because it covers multiple scenarios and expert commentary is given.
Ego-Exo4D is a large-scale video dataset with 1,286 hours of synchronized egocentric and exocentric activity videos.
It covers eight scenarios: cooking, music, soccer, health, basketball, dance, bike repair, and rock climbing. 
Various types of annotations are provided in this dataset. From these, we use \textit{expert commentary} as our source of {\textit expert knowledge}.
It contains comments on strengths, weaknesses, and actionable suggestions annotated by multiple experts for each scenario. Each comment is tied to a video segment, and some video segments have multiple comments from multiple experts.
We structure the EgoExoAsk dataset through the following procedures.

\textbf{Expert Comments Formatting.}
We start with a video $V_t$ with expert comments provided by Ego-Exo4D. The expert comments are originally tagged with ``[Good Execution]'' or ``[Tips for Improvement]''(hereafter noted as TYPE label). Since the comments are transcriptions of audio that might contain informal spoken expressions that are not optimal for training the model, we first format these comments. Besides, we also refine some comments that are too fragmented or too simple. In addition, we gather several similar comments with the same TYPE label made at the very close time stamp and determine whether these comments should be merged. This process is conducted by an LLM $\mathcal{M}(\cdot)$. Overall, we prompt the LLM to refine the expert comments in five aspects as follows:
\begin{itemize}
    \item \textit{Formalization}: reducing informal spoken expressions and meta-comments(e.g., the expert apologizing for a mistake on the previous comment);
    \item \textit{Semantic Consolidation}: merging overlapping contents into one paragraph;
    \item \textit{Categorical Clarity}: avoiding using words like ``additionally'' or ``also'' to merge comments in multiple aspects into one paragraph;
    \item \textit{Depth}: filtering out too simple or obvious comments without explanation or reasoning.
\end{itemize}

\begin{table}[t]
    \centering
    \begin{tabularx}{0.47\textwidth}{XX}
    \toprule
    Original Comments     & Formatted Comments  \\
    \midrule
``She's combining directional pulls'', ``This is wonderful'', ``She's doing this purposefully, swiftly and easily''         & The performer is effectively combining directional pulls in a purposeful, swift, and effortless manner.\\
\bottomrule
    \end{tabularx}
    \caption{An example of formatted comments.}
    \label{tab:formatted}
\end{table}
Table~\ref{tab:formatted} shows an example of a formatted comment. The full prompt and more formatted comments samples can be found in the Supplementary. After this process, we obtain comments with corresponding time stamps and TYPE labels $\{(c_{t,i}, t_{t,i}, \text{TYPE}_{t,i})|i\in [1,N_t] \}$($N_t$ being the number of comments in video $V_t$) for $V_t$.
We use Qwen3-32B~\cite{yang2025qwen3}, a powerful open-sourced LLM that can run locally to reduce the cost, for the formatting procedures. 
At last, we filter the formatted comments that are evidently short and use a small-sized LLM of another family, GPT-4o, to double-check and exclude comments merely stating an overall evaluation without explanation. These comments $c_{t,i}$ are used as the answers for $V_t$, consisting of the EgoExoAsk QA pairs.

\textbf{Question Generation.}
After the expert comments are arranged, we use the LLM, Qwen3-32B, to generate questions. The LLM generates one question $q_{t,i}$ for each formatted comment $(c_{t,i}, t_{t,i}, \text{TYPE}_{t,i})$. 
To enhance question quality, we additionally feed {\it atomic descriptions}, human-annotated captions for each video event, to the LLM. We tie all the atomic descriptions in the temporal window $[t_{t,i}-w, t_{t,i}]$ to the comment $c_{t,i}$, where $w$ is the window size.
We designed this temporal window based on the nature of Ego-Exo4D: each expert comment does not refer to events after $t_{t,i}$, the frame where the comment is made~\cite{egoexo4d}. For a sufficiently large duration of the event, we set $w$ to 7s.

Given the prompt $p^\text{q}$, the formatted comment $c_{t,i}$, and all descriptions $\mathbf{d}_{t,i}$ in the temporal window, the initial version of the question $q_{t,i}$ is generated as follows:
\begin{equation}
    q_{t,i}=\mathcal{M}(c_{t,i},\mathbf{d}_{t,i}|p^\text{q}).
\end{equation}
The full prompt can be found in the Supplementary. We design the prompt to control the LLM not to ask questions obvious in the segment. In addition, the prompt $p^\text{q}$ differs depending on TYPE labels. This is to guide the LLM to generate questions that are aware of the good/suboptimal actions in the video. We believe a model that can ask questions from different views is superior.

\textbf{Question Verification and Regenerating.}
To further improve question quality, we append a verification and regeneration process in addition to question generation. We manually list several rules that we expect the LLM to follow strictly. Given the question $q_{t,i}$, comment $c_{t,i}$ and 
captions $\mathbf{d}_{t,i}$, the LLM is asked to check the rules one by one. The rules are focused on:
\begin{itemize}
    \item Whether the question is based on observable actions(since it is expected that the question can be raised only with the video in particular);
    \item Whether the question is too general (e.g., What is important in the person's movement?);
    \item Whether the question overlaps too much with the comment (since we expect the question to be only a good trigger to the expert knowledge);
    \item Whether the question contains multiple question words (to maintain a one-to-one QA pair construction).
\end{itemize}
If the question follows all the rules, the LLM outputs ``OK'' and the question is preserved as the final version. Otherwise, the LLM outputs the reasons why the question does not follow the rules, noted as $r_{t,i}$. The final version is obtained by regeneration using the verification prompt $p^\text{v}$:
\begin{equation}
    q_{t,i}=\mathcal{M}(c_{t,i},\mathbf{d}_{t,i}|q_{t,i}, r_{t,i},p^\text{v}).
\end{equation}
$p^\text{v}$ is an extension of $p^\text{q}$, which notes the original question $q_{t,i}$ as a negative sample tells the LLM the reason. Since the $p^\text{v}$ and $p^\text{q}$ share the same main body, even though $q_{t,i}$ is falsely judged as a suboptimal question, the regeneration process can be viewed as generating an alternative question.
Note that we used Qwen3-32B again for this procedure.


\subsection{Retriever Training}\label{subsec:retriever}
After obtaining abundant QA pairs $\{(q_{t,i},c_{t,i})|i \in [1,N_t]\}_{t \in [1,N]}$ ($N$ being the number of videos), we can now train an expert-knowledge-aware retriever $\mathcal{E}$. Here, we use the videos that are originally annotated as training data in Ego-Exo4D for training. Since training $\mathcal{E}$ does not require video information, we rewrite the QA pairs as $\{(q_{i},c_{i})\}_{i \in [1,N_\text{all}]}$ ($N_\text{all}$ being the number of all QA pairs). Given the batch size $B$, the loss is calculated by reducing the distance between paired QA and extending the distance between non-paired QA, as follows:
\begin{equation}
\mathcal{L} = - \frac{1}{B} \sum_{i=1}^{B} 
    \log \frac{\exp\left(\mathrm{sim}\!\left(\mathcal{E}(q_i), \mathcal{E}(c_i)\right) / \tau \right)}
    {\sum_{j=1}^{B} \exp\left(\mathrm{sim}\!\left(\mathcal{E}(q_i), \mathcal{E}(c_j)\right) / \tau \right)},
\end{equation}
where $\tau$ is a temperature parameter, set as 0.05.

We utilized the \textit{all-MiniLM-L6-v2} pre-trained sentence transformer as the base of the QA retriever. Extracting generated 21,240 QA pairs for the training set, we trained the model with $B=512$ for 10 epochs. We used an AdamW optimizer and the linear scheduler. The initial learning rate is set to 5e-5.


\begin{table}[t]
\centering
\caption{Retrieval accuracy on the validation set of EgoExoAsk of the QA retriever before and after training. R@1,5,10 means Recall@1,5,10; MeanR is the mean rank of the retrieved ground truth.}
\label{tab:retriever}
\begin{tabular}{lcccc}
\toprule
 & R@1 & R@5 & R@10 & MeanR  \\
\midrule
Zero-shot & 0.1347 & 0.2600 & 0.3286 & 401.22\\
Trained &0.3300 &0.5512 &0.6433 &70.8 \\
\bottomrule
\end{tabular}
\end{table}

It is crucial for our evaluation method that the generated questions should not leak the answer-specific information and the trained QA retriever must be capable of reliably retrieving answers from questions.
We confirmed this through a comparison between a zero-shot and the trained retriever. If a retriever only pretrained for general documents works well on this task, answer-specific descriptions should be leaking to questions, indicating the low quality of EgoExoAsk. On the other hand, if a QA retriever trained on the training set works well, the communication between the interviewer and expert is well simulated.

Table~\ref{tab:retriever} shows the test result on the other 6,272 QA pairs not used for training. Here, the retrieval task is defined with only one positive answer for each question. Clearly, the zero-shot model, \textit{all-MiniLM-L6-v2} which is a widely-used sentence embedding model, struggles in finding the corresponding answer, indicating less answer-specific word leaks in our dataset. At the same time, the trained model works well under this severe setting: one positive answer assumption and less keyword overlap.
\section{Retrieval-based VQG Evaluation}
In this section, we introduce the retrieval-based evaluation using the EgoExoAsk and the retriever described in Sec.~\ref{sec4}. In Sec.~\ref{subsec:retrievalpool}, we describe the retrieval pool construction by attaching the formatted comments of EgoExoAsk to video segments. In Sec.~\ref{subsec:metrics}, we present the computation of retrieval metrics with the retriever.

\subsection{Retrieval Pool}
\label{subsec:retrievalpool}
\textbf{Video Segment.}
Given the video $V_t$ and the comments ${(c_{t,i}, t_{t,i})|i\in [1, N_t]}$, we group comments that share timestamp $t_{t,j}$ ($j\in [1, N_g]$; $N_g$ is the number of timestamp groups) into $\mathcal{C}^P_{t,j}$. The $j$-th segment is then clipped from the $V_{k}$ between $t_{t,j}-w$ and $t_{t,j}$, denoted as $s_{t,j}$, where $w$ is the length of the temporal window. We define that the comments in $\mathcal{C}^P_{t,j}$ are attached to $s_{t,j}$ and $\mathcal{C}^P_{t,j}$ is the positive comment group for $s_{t,j}$.

\begin{algorithm}[t]
\caption{Retrieval Pool Construction}
\label{alg:retrieval_pool}
\begin{algorithmic}[1]
\Require Positive comment set $\mathcal{C}^P_{t,j}$ for segment $s_{t,j}$, 
         video $V_t$, scenario label $\tau_k$
\Ensure Retrieval pool $\mathcal{R}_{t,j}$ of fixed size $L$

\State \textbf{Step 1. Positives:}
\State $L^P_{t,j} \gets |\mathcal{C}^P_{t,j}|$, $\mathcal{C}^N_{t,j} \gets \{\space\}_{L-L^C_{t,j}}$
\State $\text{rem} \gets L - L^P_{t,j}$

\State \textbf{Step 2. Same-video negatives:} 
\State $\mathcal{C}_{\text{vid}} \gets \{ c \mid c \text{ attached to } s_{t,v}, v \neq j \}$
\State Take $\min(|\mathcal{C}^P_{\text{vid}}|, \text{rem})$ samples from $\mathcal{C}_{\text{vid}}$ and add to $\mathcal{C}^N_{t,j}$
\State $\text{rem} \gets \text{rem} - |\mathcal{C}^N_{t,j}|$

\If{$\text{rem} > 0$}
    \State \textbf{Step 3. Same-scenario negatives:}
    \State $\mathcal{C}_{\text{task}} \gets \{ c \mid c \text{ attached to } V_m, \tau_m = \tau_t, V_m \neq V_t \}$
    \State Take $\min(|\mathcal{C}_{\text{task}}|, \text{rem})$ samples from $\mathcal{C}_{\text{task}}$ and add to $\mathcal{C}^N_{t,j}$
    \State $\text{rem} \gets \text{rem} -$ number of samples taken
\EndIf

\If{$\text{rem} > 0$}
    \State \textbf{Step 4. Random negatives:}
    \State $\mathcal{C}_{\text{others}} \gets \{ c \mid c \notin \mathcal{C}^P_{t,j} \cup \mathcal{C}^N_{t,j} \}$
    \State Take $\text{rem}$ samples from $\mathcal{C}_{\text{others}}$ and add to $\mathcal{C}^N_{t,j}$
\EndIf

\State \textbf{Step 5. Retrieval pool:}
\State $\mathcal{R}_{t,j} \gets \mathcal{C}^P_{t,j} \cup \mathcal{C}^N_{t,j}$

\end{algorithmic}
\end{algorithm}

\textbf{Retrieval Pool Construction.}
Next, we create a retrieval pool $\mathcal{R}_{t,j}$ for the video segment $s_{t,j}$. The retrieval pool consists of the positive comment group $\mathcal{C}^P_{t,j}=\{c^P_{t,j,u}|u\in [1,L^P_{t,j}]\}$ and negative comment group $\mathcal{C}^N_{t,j}=\{c^N_{t,j,u}|u\in [1,L^N_{t,j}]\}$, where $L^P_{t,j}$ and $L^N_{t,j}$ are the number of positive and negative comments, respectively. We fix the retrieval pool size $L=L^P_{t,j}+L^N_{t,j}$ across pools, while $L^P_{t,j}$ and $L^N_{t,j}$ may vary by pool.

For the negative comment group $\mathcal{C}^N_{t,j}$, we sample comments attached to other video segments. Specifically, the flow of the retrieval pool construction is summarized in Algorithm~\ref{alg:retrieval_pool}. The purpose is to prevent the retrieval task from being naive; for example, if the pool contains only one positive comment about rock climbing and negative comments about cooking, a general question about rock climbing could trivially retrieve the correct answer without targeting the specific point at issue.


\subsection{Retrieval Metrics as Question Evaluation}
\label{subsec:metrics}
Once retrieval pools are constructed for all segments, we use them to evaluate VQG's capability for eliciting expert knowledge. We adopt the widely used Recall@$k$ and rank metrics for evaluation. For segment $s_{t,j}$ with retrieval pool $\mathcal{R}_{t,j}$, the VQG model generates a question $\hat{q}_{t,j}$. We use the retriever $\mathcal{E}(\cdot)$ trained in Sec.~\ref{sec4} to compute similarities between $\hat{q}_{t,j}$ and comments in $\mathcal{R}_{t,j}$, and rank comments by similarity in descending order. If any positive comment $c\in \mathcal{C}^P_{t,j}$ is ranked top-$k$, we recognize it as a {\it good} question. Recall@$k$ is then calculated as follows:
\begin{equation}
    \text{Recall}@k = \frac{\tau_k}{\nu},
\end{equation}
where $\tau_k$ and $\nu$ are the numbers of {\it good} questions and all queries, respectively. The rank metrics are the mean and median of the best rank achieved by any $c \in \mathcal{C}^P_{t,j}$.
\section{Experiments}
\subsection{Experimental Settings}
\begin{figure*}[t]
  \centering\includegraphics[width=0.925\linewidth]{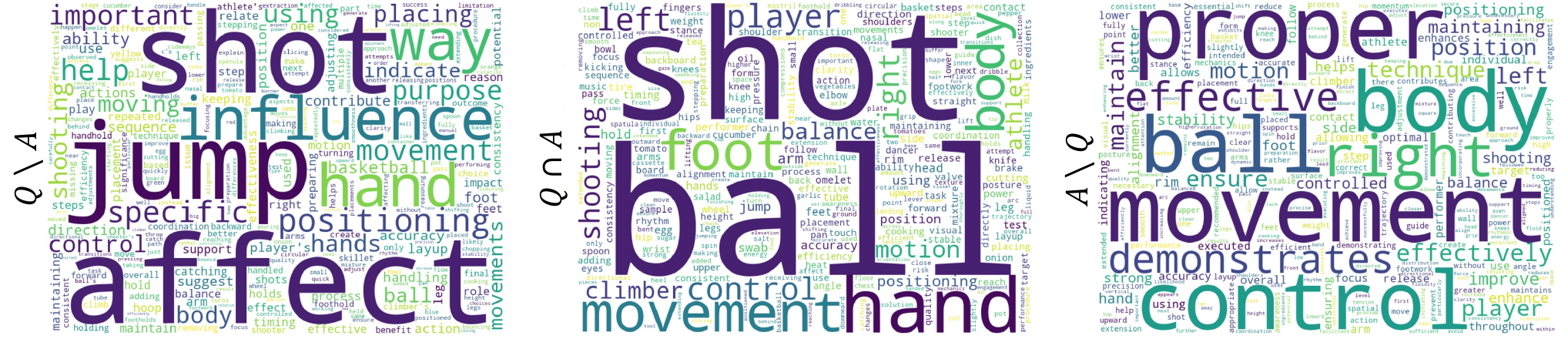}
   \caption{Word cloud for the EgoExoAsk dataset. Q $\backslash$ A shows frequent words in questions only; Q $\cap$ A in both; A $\backslash$ Q in answers only.}
   \label{fig:wordcloud}
\end{figure*}

\begin{figure}[t]
  \centering
  \includegraphics[width=\linewidth]{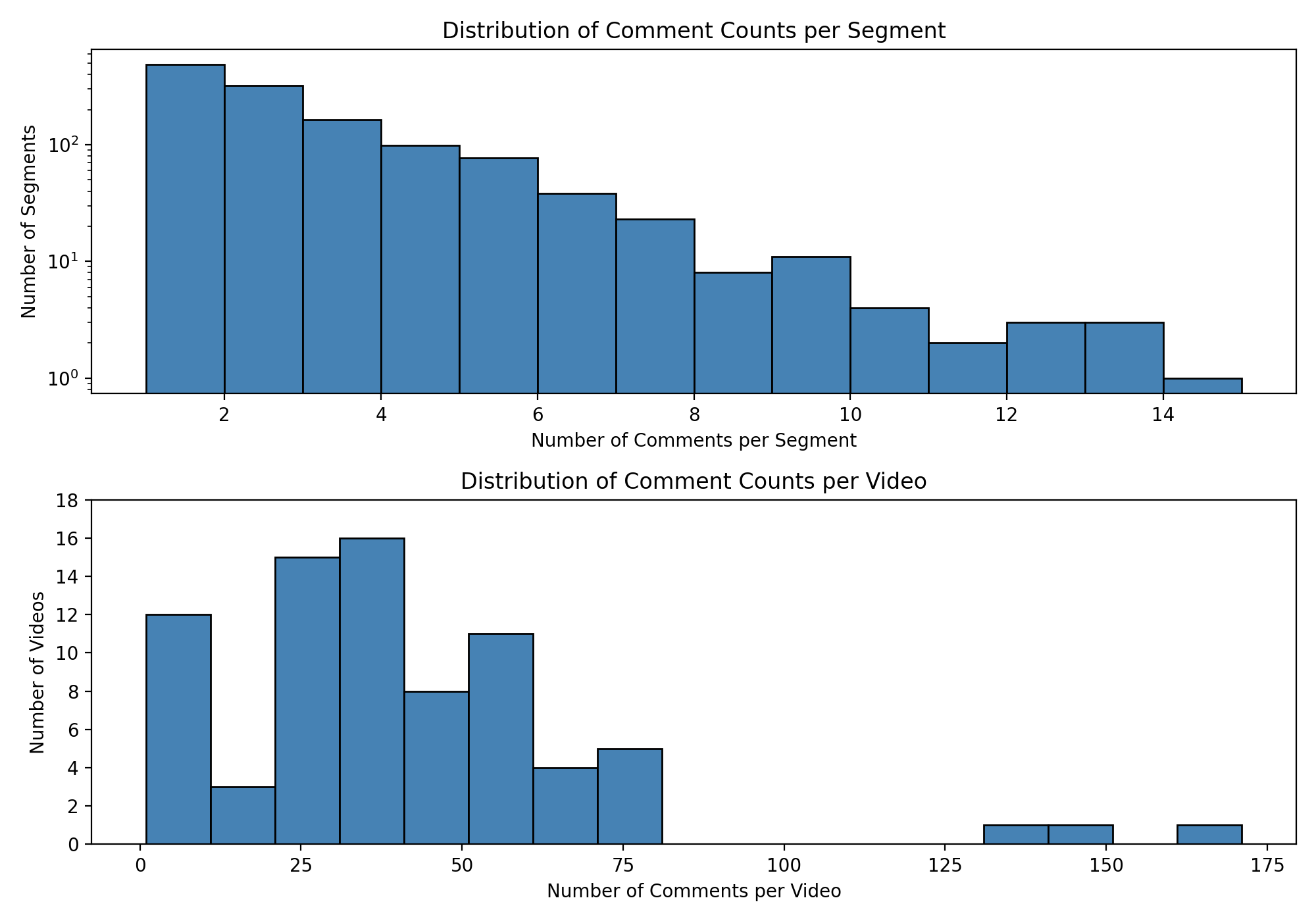}
   \caption{Histograms showing the distribution of comment number per segment and per video. For visibility, the segment histogram is shown on a log scale. Bin size is 1 for the segment and 10 for the video.}
   \label{fig:num_comments_per_video}
\end{figure}

\textbf{Benchmark}
We designed a benchmark based on our EgoExoAsk dataset. We split it into 21,240 QA pairs for QA retriever training (\ref{subsec:retriever}), 3,177 QA pairs (tied with 1,275 video segments) as seen expert knowledge for certain VQG settings, and 3,099 QA pairs (1,237 video segments) as the validation set for the VQG performance measurement.
The task for VQG models is to generate a question $q$ from an input video segment $s$ that elicits unseen knowledge behind the video from experts. Our proposed method evaluates a VQG model's performance by using the generated question to retrieve the positive comment $c^P$ with the trained QA retriever.

Ego-Exo4D provides a time-synchronized ego-centric view and multiple exo-centric views for each video. As the best exo-centric view is annotated with {\it atomic descriptions}, we selected the exo-centric view for each $V_t$ based on the annotation.

Since most segments are tied with fewer than five positive comments, and most videos contain 20 to 50 comments to provide hard negatives (shown in Fig.~\ref{fig:num_comments_per_video}), we set the retrieval pool length $L$ to 50 for this benchmark.

\textbf{VQG Settings.}
We test the capability of VQG for expert-knowledge elicitation on three state-of-the-art open-source VLLMs: Qwen2.5-VL (8B)~\cite{bai2025qwen2}, InternVideo2.5 (8B)~\cite{wang2025internvideo2} and Video-LLaVA (7B)~\cite{lin2024video}. We evaluate VQG for two view types: \textbf{exo} and \textbf{ego} videos. For each view and VLLM, three methods were applied:
\begin{itemize}
    \item \textbf{naive}: only the video is observable for the VLLM;
    \item \textbf{w/ caption}: \textit{atomic descriptions} are accessible for the VLLM in addition to the video; 
    \item \textbf{w/ RAG}: RAG provides seen expert knowledge queried by \textit{atomic descriptions}.
\end{itemize}

In practice, we employed the RAG using the \textit{all-MiniLM-L6-v2} pre-trained sentence transformer, a widely-used embedding model for RAG in text generation. The intention to use RAG is to see whether seen domain knowledge helps generate better questions.
For \textbf{w/ RAG}, we use the top-5 results retrieved with the corresponding \textit{atomic descriptions}.

We also calculated the \textbf{Gold} result as the ideal result for our benchmark. This serves as an approximate upper bound. A high \textbf{Gold} score in our framework would support the evaluation’s reliability. In this setting, we sampled a question for each video segment from EgoExoAsk as the query, instead of generating.
Since each segment has multiple questions, we randomly pick one at each retrieval. We repeated this process three times for each segment to obtain an average result.

As for the lower bound of our benchmark, we calculated the \textbf{Random} baseline. Our evaluation framework should yield higher scores for VLLMs than \textbf{Random}. The \textbf{Random} baseline is employed by randomly permuting the comments within each retrieval pool. Similar to \textbf{Gold}, we performed this procedure three times and reported the average score.
In addition, we prepared \textbf{text-only} baseline, which generates questions with \textit{atomic descriptions} of each segment instead of referring to visual data. Since the VLLMs we used have approximately 8B parameters, we used Qwen3-8B for fairness. This provides baselines for the method \textbf{w/ caption}, as the performance gain from \textbf{text-only} shows the contribution of VLLMs for the VQG task.


\subsection{Experimental Results}

\begin{table*}[t]
\centering
\caption{Evaluation results. “--” means not applicable. Bolds highlight the best performance of the model with the same FPS under different conditions, and underlines indicate the second-best.}
\label{tab:main}
\setlength{\tabcolsep}{6pt}
\scalebox{0.9}{
\begin{tabular}{l l c c c c c c}
\toprule
Method & Model & View & FPS & R@1 & R@5 & MeanR & MedianR \\
\midrule
Gold & -- & -- & -- & 0.6631 & 0.9041 & 2.61 & 1 \\
Random & -- & -- & -- &0.0501 &0.2115  & 18.26 & 15\\
\midrule\midrule
\multirow{8}{*}{\textbf{naive}} 
  & QwenVL-2.5      & \multirow{4}{*}{Exo} & 2  & 0.0614 & 0.2643 & 16.44 & 13 \\
  & QwenVL-2.5      &                       & 10 & 0.0768 & 0.2805 & 16.27 & 12 \\
  & InternVideo-2.5 &                       & 16 & 0.0695 & 0.2563 & 16.79 & 13 \\
  & Video-LLaVA     &                       & 2  & 0.0550 & 0.2142 & 18.56 & 16 \\
\cmidrule(lr){2-8}
  & QwenVL-2.5      & \multirow{4}{*}{Ego} & 2  & 0.0881 & 0.2654 & 15.43 & \underline{11} \\
  & QwenVL-2.5      &                       & 10 & 0.0914 & 0.2910 & 15.63 & 12 \\
  & InternVideo-2.5 &                       & 16 & 0.0760 & 0.2708 & 16.51 & 13 \\
  & Video-LLaVA     &                       & 2  & 0.0574 & 0.2272 & 18.16 & 15 \\
\midrule\midrule
\multirow{8}{*}{\textbf{w/ caption}}
  & Qwen3 (text-only)           & --                    & -- & 0.1043 & 0.3355 & 14.98 & 11 \\
\cmidrule(lr){2-8}
  & QwenVL-2.5      & \multirow{4}{*}{Exo} & 2  & \underline{0.0946} & \underline{0.3387} & \underline{14.72} & \textbf{10} \\
  & QwenVL-2.5      &                       & 10 & \textbf{0.1075} & \textbf{0.3226} & \underline{14.87} & \textbf{11} \\
  & InternVideo-2.5 &                       & 16 & \underline{0.1099} & \underline{0.3242} & \underline{15.43} & \underline{11} \\
  & Video-LLaVA     &                       & 2  & \underline{0.1019} & \underline{0.2805} & \underline{16.88} & \underline{13} \\
\cmidrule(lr){2-8}
  & QwenVL-2.5      & \multirow{4}{*}{Ego} & 2  & \textbf{0.1140} & \textbf{0.3395} & \textbf{14.19} & \textbf{10} \\
  & QwenVL-2.5      &                       & 10 & \textbf{0.1075} & \underline{0.3161} & \textbf{14.67} & \textbf{11} \\
  & InternVideo-2.5 &                       & 16 & \textbf{0.1124} & \textbf{0.3314} & \textbf{14.57} & \textbf{10} \\
  & Video-LLaVA     &                       & 2  & \textbf{0.1059} & \textbf{0.2967} & \textbf{16.52} & \textbf{12} \\
\midrule\midrule
\multirow{8}{*}{\textbf{w/ RAG}} 
  & QwenVL-2.5      & \multirow{4}{*}{Exo} & 2  & 0.0784 & 0.2765 & 16.63 & 13 \\
  & QwenVL-2.5      &                       & 10 & 0.0687 & 0.2700 & 16.69 & 13 \\
  & InternVideo-2.5 &                       & 16 & 0.0485 & 0.2255 & 17.33 & 14 \\
  & Video-LLaVA     &                       & 2  & 0.0841 & 0.2724 & 16.64 & 13 \\
\cmidrule(lr){2-8}
  & QwenVL-2.5      & \multirow{4}{*}{Ego} & 2  & 0.0760 & 0.2732 & 16.43 & 13 \\
  & QwenVL-2.5      &                       & 10 & 0.0719 & 0.2781 & 16.22 & 13 \\
  & InternVideo-2.5 &                       & 16 & 0.0736 & 0.2514 & 16.75 & 14 \\
  & Video-LLaVA     &                       & 2  & 0.0768 & 0.2821 & 16.62 & 12 \\
\bottomrule
\end{tabular}
}
\end{table*}
\begin{figure*}[t]
  \centering
  \includegraphics[width=0.99459\linewidth]{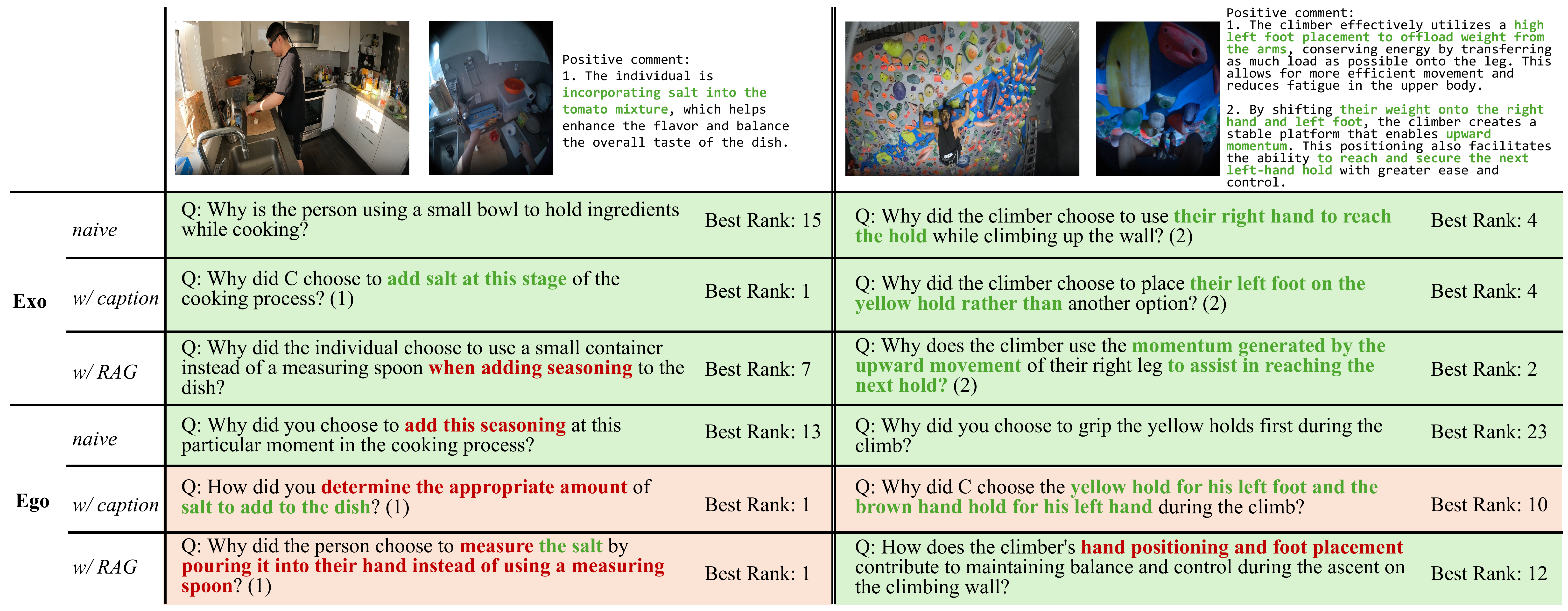}
   \caption{Qualitative results of generated questions and retrieval results. A green background represents a suitable assessment; a red background indicates that the evaluation lacks reasonability. Green-highlighted text indicates consistency between the positive comment and the question; red indicates the opposite. In addition, the label $(i)$ noted at the end of the question means it retrieves the $i$-th positive comment at the top 5.}
   \label{fig:qualitative}
\end{figure*}
\textbf{Quantitative Results.}
Evaluation results for different methods on the EgoExoAsk Benchmark are shown in Tab.~\ref{tab:main}. From Tab.~\ref{tab:main}, we first see that \textbf{Gold} achieves the highest values, which supports the evaluation's validity as \textbf{Gold} approximates oracle queries. It can also be observed that \textbf{Random} generally yields the worst results compared to other methods, serving as a reference lower bound of our benchmark.

For different VLLMs of the same size, they exhibit similar performance levels. This supports the robustness of our evaluation framework. In addition, providing video captions improves R@1 and MeanR for all VLLMs (by about 0.03 and 1.5 on average compared to \textbf{naive}, respectively). This trend aligns with the intuition that richer context helps VLLMs formulate better questions. On the other hand, the \textbf{w/ RAG} method does not always improve the metrics and sometimes can even be inferior to the \textbf{naive} method. This is likely due to the inappropriate retrieved context in the RAG process, since the retriever we used is not fine-tuned on the expert knowledge domain.

We then focus on the results of Qwen3 and QwenVL-2.5 (FPS=2). We consider the Qwen3 results because the LLM part of QwenVL-2.5 belongs to the same family as Qwen3. As shown in Tab.\ref{tab:main}, QwenVL-2.5 using ego view and atomic descriptions achieves the best scores, with R@1=0.1140 and MeanR=14.19. Since atomic descriptions capture global actions and ego videos provide egocentric cues, VLLMs with these inputs should generate more plausible questions for the segment.

FPS is another factor that we expect to affect the evaluation results. Comparing QwenVL-2.5 (\textbf{naive}) when FPS is 2 and 10, respectively, we can see that higher FPS contributes to better evaluation values from Tab.~\ref{tab:main} (R@1=0.0614 for FPS=2 and R@1=0.0768 for FPS=10). Since using higher FPS generally enables the VLLMs to capture more details~\cite{li2025improving}, the observed gains further support the reliability of our evaluation framework. On the other hand, for the \textbf{w/ caption} method, higher FPS does not yield better results. We hypothesize that additional video frames overlap with the caption, reducing information gain and increasing computational load, thereby making generation more difficult.


\textbf{Qualitative Results.}
We demonstrate some samples of questions generated by QwenVL-2.5 (FPS=2) and the corresponding retrieval results, along with the positive comments, in Fig.~\ref{fig:qualitative}.

From Fig.~\ref{fig:qualitative}, we can see that our evaluation yields reasonable results (the retrieval pool for each sample can be found in the Supplementary). In the first video sample, the question asking for \textit{``add salt at this stage''} retrieves the positive comment at the top 1, which is clearly aligned with the video segment. By contrast, questions about unimportant objects (\textit{``bowl''}) or incorrect ingredients (\textit{``seasoning''}) are assessed as suboptimal. For the second video, questions about \textit{``left foot'' and ``right hand''} retrieve the 2nd positive comment within the top 5. A more detailed question (\textit{momentum''}, \textit{upward movement''}, and \textit{``reaching the next hold''}) elevates the 2nd positive comment to the second place. In contrast, the question that only mentions generic \textit{``hand positioning and foot placement''} is not evaluated as good. These results provide intuitive verification for the reliability of our evaluation framework.

On the other hand, there are also failure cases. Even though a question retrieves a positive comment, it can be off-point (e.g., inquiring about \textit{``measuring the salt''} which is not the focus of the expert comment in the first video). In addition, for the second video, the question \textit{``Why did C choose the yellow hold for his left foot and the brown hand hold for his left hand during the climb?''} should match the 1st positive comment; however, it is retrieved only at rank 10.

\section{Limitation}
Since our retriever is trained on limited domains, further research on the generalization ability of our framework is necessary.
Also, although we design a verification-and-regeneration method to generate our EgoExoAsk dataset and avoid direct evaluation by LLM, it may not eliminate latent bias and hallucinations. Any future updates for these techniques will improve the reliability of our proposed evaluation framework. In addition, when analyzing the failures in Fig.~\ref{fig:qualitative}, we infer that they are mainly due to the suboptimal negative samples in the retrieval pool, which can be improved in future work.

Another limitation is the potential for a deeper analysis regarding whether questions specifically refer to the video content. In the proposed metric aimed at knowledge elicitation, questions may achieve high retrieval scores without explicitly referencing specific video content (e.g. \textit{``Why did you choose to climb this particular route?''}). This is not inherently contrary to our objectives, as an excellent interviewer does not always point out specific details in every question. However, by evaluating questions in terms of their specificity alongside the proposed criterion, it could be possible to conduct a more detailed analysis of effective questioning strategies. Therefore, a cross-check with human expert validation and more multi-perspective metrics should be considered in future work.

\section{Conclusion}
In this study, we focused on eliciting unseen knowledge from experts and proposed the EgoExoAsk dataset for evaluating VQG models. Along with this, we introduced a metric based on a QA retriever. Our benchmarking results demonstrated that the proposed method showed consistent trends across multiple models, indicating its reliability as a quantitative evaluation approach. 
Additionally, we quantified the performance boost provided by visual information compared to text-based question generation for the first time.

\section*{Acknowledgment}
This work was supported by JST K Program Grant Number JPMJKP25V1 and JST Moonshot R\&D Program Grant Number JPMJMS2236, Japan.

\clearpage
\appendix

\section*{Supplementary Material}
\addcontentsline{toc}{section}{Supplementary Material}

\section{Full Prompt}
\textbf{Expert Comment Formatting.}
In this prompt, \{comment\_text\} is the original expert comments in the EgoExo4D~\cite{egoexo4d}, \{scenario\_name\} is the scenario of the video, and \{task\_type\} is the TYPE label.
\begin{lstlisting}
You are an expert assistant for editing instructional commentary in skill demonstration videos. I will provide you with segments of transcribed expert commentary from videos. These comments may include informal spoken expressions (e.g., "you know...", "I guess...", "kinda", etc.) and might consist of multiple sentences that are either semantically redundant or cover different aspects of the demonstration. The video is about a {scenario_name} scenario.

#Input to be processed
{comment_text}

#Input Description:
    - Each group of comments is associated with a timestamp.
    - The comments are labeled as {task_type}.

#Output Requirements:
1. Formalization:
    - Eliminate all informal spoken language and filler expressions to produce clear, professional, and written-style language.

2. Semantic Consolidation:
    - If multiple comments convey similar or overlapping content, merge them into a single, concise paragraph.
    - If they refer to distinct points, split them into separate paragraphs, each focused on one specific point.

3. Categorical Clarity:
    - Each paragraph should address only one execution quality or improvement suggestion.
    - AVOID merging multiple suggestions into one paragraph by using transitions like "Additionally", "Also", "and", or "then".

4. Ensure Depth of Comments:
    - if the original comments are too simple or obvious, or only describe a simple action without explaining or reasoning on it (e.g., "the execution is good/effective", "she raise her left hand"), discard them.
    
5. Information Grounding:
    - DO NOT give information not mentioned in the commentary segment.
    - DO NOT add information not mentioned in the original commentary segment.

6. Ignore Meta-comments:
    - If any comment appears to reflect on the annotation process itself (e.g., apologizing for an error, noting that the expert did not fully review the video, expressing confusion or excitement about the video, stating invisibility of the video), do not include it in the output, even if it is phrased in an instructional tone. Only keep comments that clearly relate to the physical performance or improvement of the task shown in the video.

7. Labeling:
    - Each output segment should be labeled by [Good Execution] or [Tip for Improvement] according to the original label.

#Output Format:
[<Label>] comment1
[<Label>] comment2
...
\end{lstlisting}

\textbf{Question Generation.}
We give different prompts about questions labeled with ``[Good Execution]'' and ``[Tips for improvement]'' as follows:
\begin{lstlisting}
TEMPLATE_QUESTIONTYPE = {
"good_executions": f"""The aim of the question is to uncover the notifiable actions or techniques.
    - If the Expert Commentary gives reasons for why the execution is good or methods of how to conduct this good execution, the question should lead to these reasons or methods based on the observed execution.
    - If the Expert Commentary describes how the execution is well conducted, the question should lead to these descriptions.""",
"tips_for_improvement": f"""The aim of the question is to uncover the actions that are not optimal.
    - If the Expert Commentary gives a suggestion to improve the execution, the question should lead to these suggestions based on the observed execution, not asking questions as if you know the "Expert Commentary".
    - If the Expert Commentary describes how the execution is poorly conducted, the question should lead to these descriptions.""",
}
\end{lstlisting}

In the prompt, \{desc\_text\} is the atomic descriptions of the segment, \{A\} is a formatted comment, and \{goal\_template\} is selected based on the TYPE label.
\begin{lstlisting}
You are an AI assistant tasked with generating insightful questions. You will be observing a task video, presented as a "Narration" (seen as the surface-level actions) and the corresponding "Expert Commentary" (seen as deeper insights, reasons, or techniques). 

#Narration of the video (seen as the surface-level actions)
{desc_text}

#Expert commentary of the video (seen as deeper insights, reasons, or techniques)
{A}

#About the video
This is a video of {scenario_name} scenario.

#Goal
Your goal is to formulate questions specific on the observed scene after seeing the "Narration" to understand the deeper insights revealed in the "Expert Commentary", keeping in mind the specific context provided above. The questions should ultimately guide the learner towards an understanding similar to what an expert might articulate. The Expert Commentary is labeled as {task_type}.

#Guidelines

##Should
    1. Ask for more than what is obvious from the "Narration" alone. It should probe into the *reasons*, *intentions*, *subtle techniques*, or *critical judgments* highlighted or implied by the "Expert Commentary".
    2. The question should be conceptually related to the "Expert Commentary", but not presuppose the "Expert Commentary".
    3. Based on the observed actions.
    4. {goal_template}
    
##Should NOT
    1. DO NOT ask for general evaluations, summaries, or opinions, such as:
        - "What is the overall quality of the performance?" (or other semantically similar phrases)
        - "What are the main points of the expert commentary?" (or other semantically similar phrases)
    2. AVOID presuppose a *positive or negative* relationship between two observed actions.
    3. AVOID presuppose a *positive or negative* consequences of the actions.
    4. SHOULD NOT be TOO detailed by using technical terms overlap with the "Expert Commentary"

##Word Restrictions
    1. AVOID vague phrases like "the video", "this movement", or "the performance".
    2. DO NOT use words like "as noted/mentioned in the expert commentary" or other similar phrases that refer to the Expert Commentary.
    3. DO NOT mention the timestamp of the video in the question.
    4. DO NOT use multiple question words in one question (such as "What ..., and how...?")
    5. AVOID questions that are too general, contain words like:
        - "specific", "aspect", "positioning", "movement", "contribute", "help", "improve", "stability", "effectiveness", "adjust"...

##Information Grounding
    1. DO NOT give information or ask about information not mentioned in the provided "Narration" and "Expert Commentary".

##Output format (there must be a prefix "[question]")
[question] ...
\end{lstlisting}

\textbf{Question Verification.}
In this prompt, \{Qe\} is the initial version of the generated question.
\begin{lstlisting}
    Verify whether the generated question. The question is generated after observing a task video, presented as a "Narration" (seen as the surface-level actions) and the corresponding "Expert Commentary" (seen as deeper insights, reasons, or techniques). The question is supposed to be formulated after seeing the video to understand gain the information in the "Expert Commentary".
            
If the question follows the rules, output "OK". If not, give reasons. When generating the reason, do not mention the index of rule that is violated.

#Question
{Qe}

#Expert Commentary
{A}

#Narration
{desc_text}

#Rules
Check the rules one by one.
    1. based on the observable actions
    2. not asking for overall evaluations, summaries, opinions or suggestions
    3. not using technical terms that overlap with the "Expert Commentary"
    4. not using multiple question words in one question

#Output format
- If the question follows the rules, only ouput:
OK

- If not, output the reasons:
Reason: ____
\end{lstlisting}

\textbf{Question Regeneration.}
This prompt is constructed by adding the following text to the \textbf{Question Generation} prompt. \{reason\} is the reason generated by the Question Verification process.
\begin{lstlisting}
##Bad Example and why it is bad (DO NOT generate question like this by avoiding the reason below)
[question] {Qe}
[reason for why it is bad question] {reason}
\end{lstlisting}

\begin{table*}[t]
\centering
\caption{EgoExoAsk Samples.}
\label{tab:egoexoasksample}
\begin{tabularx}{\textwidth}{lXX}
\toprule
Scenario & Question & Comment  \\
\midrule
cooking &What might be missing in the selection and placement of items that could affect precision in the task? & [Tip for Improvement] Gather the remaining tools, including a spoon and a measuring cup or measuring spoon, to measure the sweetener accurately. \\
\midrule
music &  What role does the timing of finger placement play in achieving consistent dynamics and articulation while playing the guitar? &[Good Execution] The guitarist demonstrates efficient technique by placing the lower finger down in a timely manner, which contributes to an excellent display of uniformity in both dynamics and articulation. \\
\midrule
soccer &What specific body positioning allows the player to stop and control the ball effectively with one leg? &[Good Execution] The player maintains a stable and effective body position, with the foot in a proper alignment, toe pointed downward and knee positioned over the ball. This posture supports control and balance, enabling precise execution of the action.\\
\midrule
health &What might be the purpose of tapping the mannequin and then removing the hands before repeating the action? &[Good Execution] The responder demonstrates effective initial assessment by physically tapping the victim and asking questions to determine their responsiveness. This approach is critical for evaluating the victim's condition and ensuring an appropriate response based on their level of consciousness.\\
\midrule
basketball &What role does bending the knees play in how the ball is released during the throw? &[Good Execution] The player effectively loads into her legs by squatting down, which allows the ball to gain lift upon release. This positioning contributes to a strong follow-through and proper placement of the shooting hand underneath the ball.\\
\midrule
dance &How might the lack of movement in the arms affect the overall expression and flow of the body during the sequence of steps and jumps? &[Tip for Improvement] Her arms remain rigidly at her sides, contributing to a static and uninvolved upper body presence. A more expressive use of the arms would enhance engagement and articulation of movement.\\
\midrule
bike repair &Why is it important to break the tire bead free from the rim before moving the wheel? &[Good Execution] Once the air has been fully released, the mechanic ensures that the tire bead is broken free from the rim. This allows for easier removal of the tire. \\
\midrule
rock climbing &How might the climber's use of the right foot affect his ability to maintain a relaxed grip on the holds? &[Tip for Improvement] The climber's current positioning limits his ability to relax on the hold due to poor opposition in his posture. It would be beneficial to observe how much force is being applied by the right foot, as this could indicate whether adjustments in lower-body positioning might improve balance and reduce upper-body tension.\\
\bottomrule
\end{tabularx}
\end{table*}

\section{EgoExoAsk Sample}
We show some EgoExoAsk samples of QA (question-comment) pairs in Tab.~\ref{tab:egoexoasksample}.

\section{Formatted Sample}
We show some more formatted comments samples of QA (question-comment) pairs in Tab.~\ref{tab:moreformatted}. The original comments come from EgoExo4D.

\begin{table*}[t]
    \centering
    \begin{tabularx}{\textwidth}{p{8cm}X}
    \toprule
    Original Comments     & Formatted Comments  \\
    \midrule
\multirow{2}{*}{\makecell[{{p{8cm}}}]{"she's doing a good job by holding on to the tube or buffer system with one hand pretty secured", "She's got her swab stick inside and she's stirring it around in there", "ensuring that she's getting every last bit or, you know, all that she can of her specimen off of the tip of the swab and into the liquid here", "She's doing a really good job of doing that."}} & 1. The individual is effectively using one hand to securely hold the tube or buffer system. This ensures stability during the transfer process.\\
& 2. The swab stick is being inserted into the liquid and moved in a manner that promotes thorough mixing. This technique helps maximize the collection of specimen material from the swab tip and facilitates its transfer into the liquid medium.\\
\midrule
\multirow{2}{*}{\makecell[{{p{8cm}}}]{"I like all the prep that's going into it.", "She is not only looking at the hold, she's also brushing the hold.", "this climber is showing that she has an understanding of what that means by brushing the hold and she wants to have better grip with these handholds", "This climber starts by brushing the holds."}} & 1. The climber demonstrates a clear understanding of technique by brushing the holds before grasping them. This action indicates an awareness of improving grip quality and preparing for secure contact with the handholds.\\
& 2. In addition to brushing, the climber also carefully inspects the hold, showing attention to detail and preparation prior to engaging with the climbing surface.\\
\midrule
\multirow{2}{*}{\makecell[{{p{8cm}}}]{"we'd love to see a big one-two step to gather momentum", "Take a little bit off when hitting the backboard.", "Want the ball to hit the backboard as it's going down, not as it's going up.", "Follow the fundamentals."}} & 1. Execute a strong one-two step to generate momentum before contacting the backboard. Time the movement so that the ball strikes the backboard while descending, not ascending.\\
& 2. Maintain proper fundamentals throughout the motion.\\
\bottomrule
    \end{tabularx}
    \caption{More examples of formatted comments.}
    \label{tab:moreformatted}
\end{table*}

\newpage~
\newpage
\section{Retrieval Pool for Qualitative Results}
The retrieval pool constructed for the first video is as follows:
\begin{lstlisting}
The root end of the green onion is properly removed.
The preparer is making cross cuts to create small rounds of the green part of the onion. This technique demonstrates a clear understanding of how to achieve uniformity in vegetable cutting.
A Japanese or Chinese cleaver is being used for the cutting task, which is an appropriate tool choice for this type of vegetable preparation.
The cutting is taking place on a wooden cutting board, which is suitable for use with heavy cleavers and helps preserve blade sharpness.
The preparer maintains proper hand positioning: knuckles are out, fingertips are tucked under, and fingers are pressing down to stabilize the vegetable during the cutting process.
The cutting motion is efficient, indicating good control over the knife and a steady rhythm that contributes to consistent results.
The individual is correctly preparing the garlic by carefully removing the cloves from the bulb. This step ensures that each clove can be used separately in cooking and helps prevent the garlic from burning due to uneven sizing.
The chef is using the flat side of the knife to press down on a clove of garlic, which helps break the skin and facilitates its removal. This technique also aids in releasing the garlic's natural oils.
Remove the remaining portion of the white paper.
The chef maintains proper knife control by anchoring the tip of the blade on the cutting board with his left hand, allowing his right hand to perform the cutting motion effectively.
He positions the garlic cloves close together and applies a consistent up-and-down cutting motion with the knife to mince all three cloves simultaneously.
The individual cracks the eggs directly onto the countertop corner before transferring them into the bowl, demonstrating a smooth and efficient workflow.
Whisking effectively combines the yolk and white, ensuring a homogeneous mixture. Adding salt at this stage helps to evenly distribute flavor throughout the ingredients.
The cook effectively manages the timing by allowing the pot to warm up while preparing the egg mixture, ensuring that both components are ready in coordination with one another.
The cook is using a nonstick skillet placed over the heat source of a gas stove. The gas flame has been properly ignited and positioned beneath the skillet to begin heating it.
A saute pan has been selected and placed on a heat source to preheat. A neutral-flavored oil is being added to the pan. The use of a neutral-flavored oil helps create a nonstick surface, which is ideal for preventing food from adhering during cooking. Examples of suitable oils include canola oil and grapeseed oil.
Additional oil is being introduced into the pan to ensure an even coating and optimal heat distribution.
Swirling the pan allows the oil to spread evenly across the base, ensuring uniform coating for consistent cooking results.
As the pan is heated, another swirl is made to ensure the coating is applied more evenly across the base.
The oil will have reduced viscosity.
The egg is introduced into the pan without producing a significant sizzling sound, indicating that the pan may not have reached an optimal temperature for searing.
To ensure even distribution of salt, it is recommended to briefly whisk the ingredients again before proceeding. This helps prevent the salt from settling at the bottom of the dish.
The chef begins swirling the egg mixture in the pan, using chopsticks to pull the cooked edges inward. This technique allows uncooked portions of the egg to spread into the available space, promoting even cooking and a consistent texture.
To ensure the egg cooks evenly, allow the uncooked portion to spread into the available space while continuing the cooking process.
The pan was used consistently throughout the process, and the movement of onion and garlic into the pan was executed smoothly.
The preparer is using a pancake turner to gently press down on the tomatoes in order to extract some of their juices. This technique helps facilitate the release of liquid, which can be beneficial for subsequent steps in food preparation.
The juice content of tomatoes can vary significantly depending on their size and ripeness. Larger, fully ripe tomatoes tend to release more liquid when cut into quarters or wedges. In contrast, smaller varieties such as cherry or grape tomatoes generally contain less moisture and will not yield as much liquid when prepared in a similar manner.
The individual is effectively using heat to steam the mixture, which helps in evenly distributing warmth and enhancing the blending process.
The technique effectively extracts both the juices and the flavor from the tomatoes, which contributes to creating a rich sauce for the eggs.
The individual is incorporating tomato sauce into the mixture, which contributes additional depth and flavor to the sauce being prepared.
Pureed tomatoes are also being added to the mixture, further enriching its base and consistency.
The execution enhances the tomato flavor by carefully balancing acidity and sweetness, resulting in a more vibrant and well-rounded taste profile.
During the off-season, when tomatoes may not be at their peak quality, smaller tomatoes will contribute less of the rich, robust tomato flavor typically desired in such dishes.
Apply pressure to the tomatoes to achieve the desired effect.
The individual is incorporating salt into the tomato mixture, which helps enhance the flavor and balance the overall taste of the dish.
The use of freshly ground black pepper effectively enhances the flavor profile by introducing depth and complexity to the dish.
The seasoning was applied thoroughly and evenly, ensuring optimal flavor distribution.
The cooked scrambled egg is incorporated into the tomato mixture, where it is gently broken down into smaller pieces and heated to ensure it reaches a uniform temperature with the sauce.
The visual appearance of the eggs indicates a successful execution at this stage of preparation.
The execution was effective, as the egg was not overcooked.
Using a smaller, more precise knife would enhance the accuracy and control when chopping the green onion.
Ensure all ingredients are organized and prepared prior to beginning the cooking process to maintain efficiency and clarity during execution.
Whisking should be performed using a consistent bottom-to-top circular motion to effectively incorporate air into the mixture, resulting in a fluffy egg texture. The use of appropriate tools such as a fork or wire whisk is recommended to facilitate this process. The objective is to achieve a fully homogeneous mixture where no streaks of egg white or yolk remain visible.
The skillet should have been allowed to heat sufficiently before adding the egg mixture. A proper sizzle upon contact indicates that the pan has reached an appropriate temperature, which is essential for achieving a well-cooked result.
The cook may have improved the dish by sauteing the garlic and tomatoes before proceeding.
It will be necessary to remove the egg from the pan and allow it to rest temporarily.
Allowing the egg to sit and cool before reintroducing it to the hot mixture increases the risk of overcooking, as the residual heat from the surrounding ingredients will continue to cook it further. To maintain optimal texture, the egg should be incorporated into the mixture while still warm but not fully cooled.
The item should be transferred to a separate container before refrigeration. It is recommended to use a glass container with a properly fitting lid to ensure safe and appropriate storage.
Salt can be used to counteract bitterness in tomatoes and also enhances the overall flavor of the dish. If bitterness is present, salt is an effective option; sugar may also be used as an alternative.
A significant portion of the liquid had evaporated during the process.
\end{lstlisting}
The retrieval pool for the second video is as follows:
\begin{lstlisting}
The climber begins the attempt by thoroughly brushing the holds, particularly the older ones, to remove excess chalk. This ensures a clean surface for better grip and optimal performance.
Brushing the holds is a strategic technique that provides climbers with an advantage by allowing them to clean the surface and restore friction. This action also enables a more thorough visual and tactile assessment of the hold, which can be crucial for planning movement and improving grip confidence.
The act of brushing facilitates a closer inspection of the hold's texture and shape, offering insights that may not be apparent from a distance. This investigative approach helps climbers make informed decisions about their next moves and enhances their overall route-reading ability.
The right hand is positioned in the pocket, and both the left and right feet are firmly planted against the wall.
The climber takes a deep breath to regulate their breathing in accordance with the intensity and pace of the climb.
The climber maintains an active grip on the starting holds, demonstrating focus by visually identifying the next hold before making a significant leftward movement to reach it.
The selected hold is of sufficient size and quality to support a down-pulling position, indicating good route-reading and hold assessment.
There is an effective use of a thumb catch underneath the hold, which enhances grip security and engagement.
The performer promptly retracts the limb upward, maintaining a high left foot position to effectively align with the lower section of the Waco Jug.
The climber maintains a straight right arm while in a tethered position, which contributes to stability and control. This allows for an effective movement sequence as the climber transitions through a slanted hole to establish a more advantageous position on the right side.
The decision to cross under with the hands demonstrates strategic thinking and planning ahead, enabling the climber to progress upward and to the right in a fluid and efficient manner.
The footwork during this movement is well-executed, supporting the climber's balance and positioning throughout the transition.
In this position, the climber is primarily supported by the right hand, illustrating how one strong anchor can be used effectively to manage body weight and momentum.
The climber is using excessive upper body strength to reach the next hold, which is inefficient and counterproductive. This approach consumes significant energy and works against the natural swing of the body. To execute such a move more effectively, it is essential to lock in the feet and utilize body positioning to maintain balance and control during the movement.
The climber effectively uses a heel hook by transferring their body weight upward, allowing them to elevate their feet and reduce the load on their hands. This technique enables the climber to maintain a higher foot position while using the mechanical advantage of the heel to pull the body closer to the wall.
The placement of the left heel is precise, with the heel and toes pointing outward. This positioning facilitates a stable hold that allows the climber to twist their foot, creating a camming action between the heel and toe against both the upper and lower surfaces of the mouth-shaped hold. This mechanical engagement significantly reduces the reliance on hand strength.
By bringing the feet into a higher position, the climber utilizes the heel-toe cam as a stable base. This position locks down the left hand and provides a solid foundation for transitioning out of the current cross and rolling over to the next hold.
The climber demonstrates effective use of the right foot by applying strong pressure to the starting hold, which supports the upward movement and contributes to maintaining balance during the transition.
The execution of the action is effective and demonstrates a high level of proficiency.
The athlete should allow a slight sag in the left arm to create balance, while maintaining full extension in both arms. This positioning will provide a stable and efficient posture, enabling smooth foot movement without resistance or unnecessary effort.
The climber effectively executes a technical movement by combining a left heel hook with a right-hand lock-off. This coordinated effort generates upward and inward force, enabling the climber to progress their body position closer to the wall. The integration of these two elements allows for a smooth transition into securing the next hold with the left hand.
After establishing the right-hand lock-off, the climber drops the right foot while maintaining heel-toe camming, which facilitates progression of the left hand up the wall. Weight is then transferred between the right hand and left foot, allowing for a controlled shift of the hips to the right. This creates a stable platform from which the climber can continue ascending.
The climber releases the left heel from the yellow hold in a fluid manner, allowing momentum to carry the leg over to the opposite side of the wall. This approach avoids resisting the natural motion and enables immediate placement of the foot on the new surface. Such a technique is particularly effective when both feet are required to leave contact with the wall simultaneously.
The climber effectively utilizes momentum to control the swing and elevate their right foot onto the foothold. This technique demonstrates efficient energy transfer, allowing them to maintain balance and progress upward.
With the left hand anchored, the climber applies downward force through the right foot, generating significant power. This positioning enables a stable and effective movement into a favorable climbing position.
The climber initially contacts the hold with an open hand and extended fingers, which is a more relaxed and energy-efficient grip. Maintaining this open-handed contact when possible can help conserve physical energy during the climb.
Gradually shift the individual's weight while maintaining a wide stance. Transfer the weight to the right side to establish a new tension point and use the right hand as an anchor.
Elevate the left foot to enhance positioning and attempt again with increased leverage over the hold.
Be mindful that this type of grip can lead to rapid fatigue if frequently used during holds.
The climber effectively utilizes a high left foot placement to offload weight from the arms, conserving energy by transferring as much load as possible onto the leg. This allows for more efficient movement and reduces fatigue in the upper body.
By shifting their weight onto the right hand and left foot, the climber creates a stable platform that enables upward momentum. This positioning also facilitates the ability to reach and secure the next left-hand hold with greater ease and control.
The use of a stick brush allows for efficient cleaning at higher levels without the need to climb, thereby conserving energy and improving access to elevated areas.
Plan the placement of brush strokes based on the intended use of each hold. Position yourself as close as possible to the next hold before committing to a move, in order to maintain efficiency and reduce unnecessary movement.
Brushing the holds before an attempt helps remove excess chalk, which enhances grip quality by improving the contact between the hands and the handhold.
The climber is using a grip that appears awkward and inefficient, with the fingers excessively extended and the thumb positioned in an unusual way on the underside of the hold. This grip seems to require unnecessary bending of the elbow and wrist, which may compromise control and stability.
A more effective approach would involve using a thumb catch on the hold. This technique provides better opposition between the thumb and the remaining four fingers, allowing for improved control and the ability to manipulate the hips into a higher body position. This adjustment can enhance overall efficiency and reduce strain during the movement.
The climber's current positioning limits his ability to relax on the hold due to poor opposition in his posture. It would be beneficial to observe how much force is being applied by the right foot, as this could indicate whether adjustments in lower-body positioning might improve balance and reduce upper-body tension.
The climber is not fully transferring body weight onto the left foot before moving the right hand. A more complete weight shift to the left foot would provide greater stability and reduce the need for excessive upper-body effort when reaching for the next hold.
The movement is being executed too quickly, with insufficient extension of the right leg inside the frame. Extending the right foot further outward would allow for a more controlled and deliberate transfer of the right hand to the higher hold.
The climber catches the hold with significant force, which is inefficient at this early stage of the climb. High-effort movements at the beginning can deplete energy reserves and hinder the ability to complete the route effectively.
The climber is over-relying on upper-body strength rather than utilizing lower-body power and balance. A more effective approach would involve driving down through the feet and maintaining an opposite-arm-and-leg positioning to stay engaged with the wall.
If the climber had placed the right foot accurately before moving the right hand up, it might have created a small but destabilizing moment off the wall, requiring increased engagement from the arms and core to re-establish contact with the wall. Proper foot placement could help avoid such instability.
The climber should adjust the placement of their right foot to ensure better stability and control, as improper positioning increases the risk of losing contact with the wall. A more secure foothold will help maintain balance during movement.
To improve efficiency, the climber should open their body position further. This adjustment will create better reach and allow the right hand to transition smoothly to the next hold.
Dropping the right foot and shifting weight forward enhances stability. To counterbalance the downward force, engage by pulling upward with the left heel toward the right. However, the practitioner does not effectively transition out of the heel hook position. His movement results in his feet being cut off, placing excessive reliance on his left and right arms to support his weight.
The athlete should focus on driving the knee over the opponent's foot to maximize weight transfer away from the hands, allowing for greater efficiency and control in the movement.
Climbers should avoid placing excessive weight on a single foot, particularly when the foot is positioned between the hands in a bent-arm position. In this case, the climber placed too much reliance on the left foot, which was not optimally placed. As a result, when the foot lost contact with the hold, the climber struggled to maintain balance and control. This situation increases the risk of finger injury due to sudden eccentric loading and overextension of the fingers.
Proper foot placement is critical for maintaining stability and efficiency during climbing movements. The climber did not carefully assess or adjust the positioning of their left foot during the transition from right to left hand holds. This lack of attention led to an unstable foothold, which contributed to the loss of control. Climbers must focus intently on where their feet are placed and ensure that they are in a secure and functional position before committing to the next move.
A sudden loss of foot contact can make it extremely difficult to recover unless the climber has exceptional strength and body control. In this scenario, the climber attempted to compensate by shifting their left hand to a new hold, but the repeated loss of the left foot disrupted their momentum and balance. This highlights the importance of ensuring that each foot placement is deliberate and stable before progressing through the movement.
Climbers should be fully aware of how their feet contribute to pushing and pulling forces against the wall. Effective climbing requires the ability to stand actively through the toes, using the feet to support body weight and generate upward force. Neglecting this aspect can lead to inefficient movement patterns and increased strain on the upper body.
\end{lstlisting}

{
    \small
    \bibliographystyle{ieeenat_fullname}
    \bibliography{main}
}

\end{document}